\documentclass[conference]{IEEEtran}
\IEEEoverridecommandlockouts    

\usepackage{multirow}
\usepackage{lipsum}
\usepackage{amsmath}
\usepackage{amssymb}
\usepackage[ruled,vlined]{algorithm2e}
\usepackage{graphicx}
\usepackage{url}
\usepackage{cite}
\usepackage{tikz}
\usepackage{wrapfig}
\graphicspath{{./Figures/}}

\title{\LARGE \bf Emergent Autonomous Drifting for Collision Avoidance in Real-World Winter Driving Scenarios}

\author{
	\parbox{\textwidth}{%
		\centering
		Elliot Weiss, Michael Thompson, Thomas Lew, and John Subosits
	}%
	\thanks{The authors are with Toyota Research Institute. Corresponding email:
		{\tt elliot.weiss@tri.global}.}%
}

\begin{document}
	
\maketitle
\thispagestyle{empty}
\pagestyle{empty}

\begin{abstract}
Real-world collision avoidance is a core motivation for studying the dynamics and control of high sideslip drifting in vehicles, yet the practical benefit of such maneuvers has so far primarily been tested in scenarios explicitly engineered to require drifting. In this work, we explore the question of if and when drifting may be optimal for safety in real-world winter driving conditions. We present a drift-capable nonlinear model predictive control (MPC) system designed to handle scenarios grounded in crash fatality data and deploy the controller in a high fidelity simulator across road departure and oncoming vehicle collision avoidance scenarios. The controller naturally initiates and sustains drifting maneuvers to stay on the road when hitting a patch of ice on the rear axle and to avoid an oncoming vehicle that has slid into its lane. Comparisons with a benchmark electronic stability control (ESC) system demonstrate how a drift-capable controller can trade off stability for controllability to precisely maneuver through dangerous winter driving scenarios. A Monte Carlo study over random ice patches further shows that the drift-capable controller achieves lower median lane error than ESC across several speeds, while revealing that drifting emerges predominantly at higher speeds.
\end{abstract}

\section{Introduction}
\label{sec:introduction}

Researchers have studied the dynamics and control of vehicles performing high sideslip drifting maneuvers for many years with the motivation of improving road safety. Even though it may be counterintuitive for most drivers to operate their vehicle in drifting regimes characterized by fully sliding tires and open-loop unstable dynamics, many works in drifting literature claim that these types of maneuvers may be necessary in emergency collision avoidance scenarios \cite{velenis2010steady, hindiyeh2013dynamics, goh2016simultaneous, chen2023dynamic, djeumou2025reference}. Collision prevention on low friction surfaces such as snow and ice, in particular, is often emphasized as a primary use case for drifting in the real world \cite{ono1998bifurcation, falcone2007predictive, jia2024novel, djeumou2024one}. Our work aims to directly explore this claim by answering the following open questions: \textit{When, if at all, is drifting the best way to avoid a collision in dangerous real-world scenarios? How does a controller that allows drifting as needed for safety (which we call ``drift-capable" control) compare to stability control technology that actively prevents drifting?} 

Although drifting research has typically focused on the task of tracking specific drift equilibria and high sideslip trajectories in race track settings, recent work has begun studying controllers that use oversteering drift maneuvers to maintain safety in scenarios modeled after more realistic driving environments. For example, Zhao \textit{et al.} present a switching mode controller that drifts to avoid a sudden pop-up obstacle \cite{zhao2021collision}, justifying its potential deployment as an active safety system through backward reachability analysis \cite{zhao2022justifying}. Autonomous drifting in pop-up obstacle scenarios has also been considered with reinforcement learning-based controllers \cite{zhao2024autonomous, liu2025deep}. In addition, Li \textit{et al.} develop a hierarchical control system with offline planning that executes a $180^{\circ}$ drift to avoid two vehicles closing down its driveable space \cite{li2023planning}, while other researchers have examined emergency lane changes due to stopped vehicles requiring drifting maneuvers \cite{zhao2024auto, wu2025mppi, li2025human} and drifting control for collision avoidance while making tight left-hand turns at intersections \cite{stano2023enhanced}. 

\begin{figure}[t]
  \centering
  \includegraphics[width=\linewidth, trim=0cm 3cm 0cm 3cm, clip]{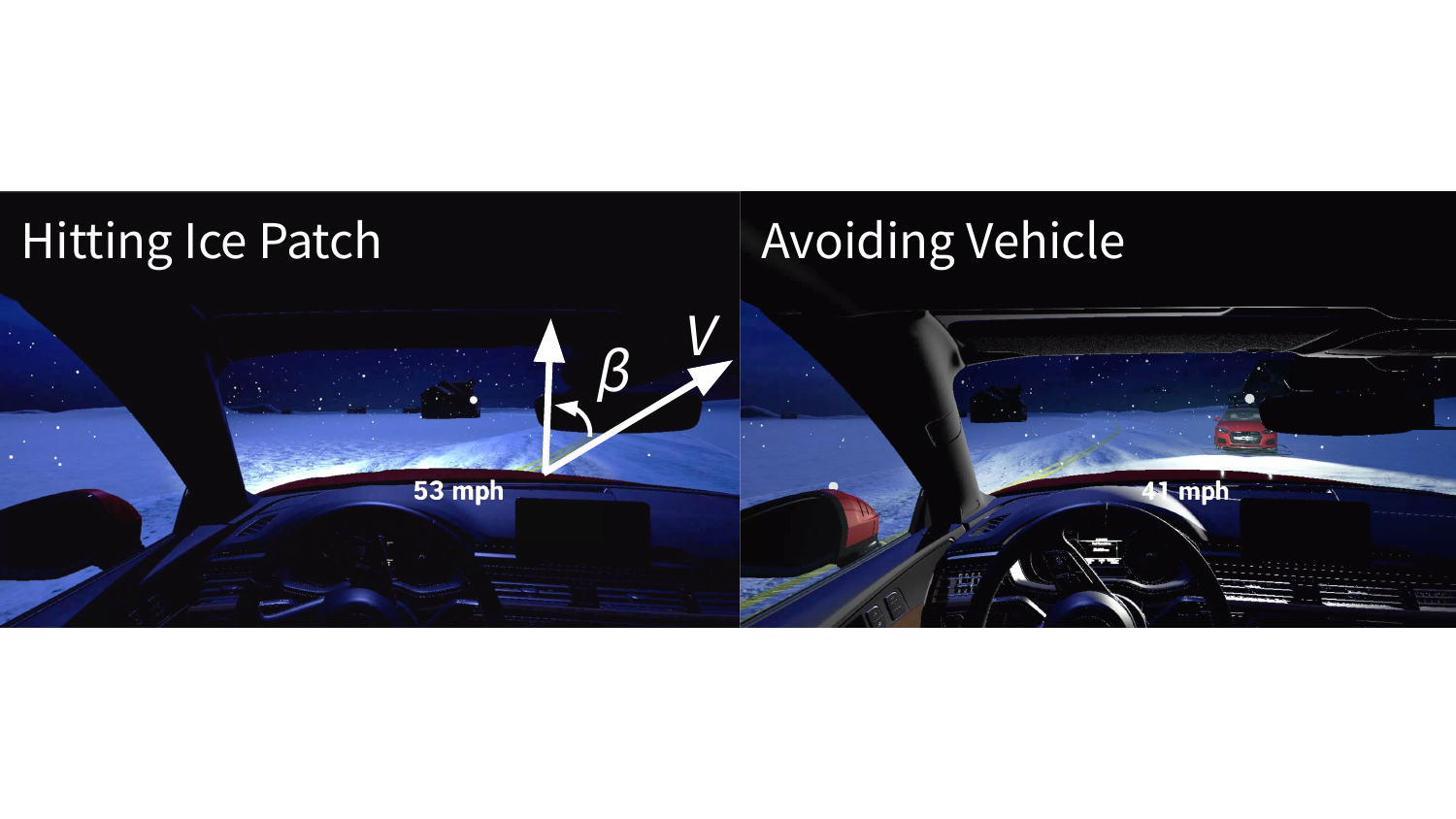}
  \caption{The proposed drift-capable controller can safely handle an ice patch on the road with a high sideslip $\beta$ maneuver (left) and avoid a collision with an oncoming vehicle (right).}
  \label{fig_teaser}
\end{figure}

This emerging body of work on drifting for vehicle safety has so far relied on scenarios explicitly designed to require drifting, such as pop-up obstacles that cannot be avoided with grip driving \cite{zhao2022justifying}, and/or controllers with complex multi-layered architectures that track pre-computed drifting trajectories \cite{li2023planning, wu2025mppi}. To assess the claim of drifting as a useful maneuver for improving road safety, our work studies \textit{emergent drifting behavior in scenarios extracted from real-world driving data}, two examples of which are shown in Fig. \ref{fig_teaser}. We specifically focus on winter conditions, given the high propensity of control-loss related incidents on low friction surfaces due to limited and varying tire-road friction \cite{weiss2025mixed}. Only a few previous studies on drifting for collision avoidance have tested controllers in winter conditions \cite{wu2025mppi, fu2024real}, leaving the broader space of high sideslip driving in realistic adverse weather scenarios largely unexplored.

\textbf{Contributions:} We present a nonlinear model predictive control (MPC) system to evaluate how and when drifting emerges as the optimal strategy to stay safe in real-world winter scenarios. Specifically, our work contributes the following:
\begin{itemize}
    \item We construct a realistic set of dangerous winter driving scenarios (speed reduction around curves, series of ice patches, imminent collision with oncoming vehicle) grounded in crash data and road design standards.
    \item Using an intuitive drift-capable MPC controller with no grip-drift mode switching or explicit drift trajectory tracking, we show that high sideslip maneuvers emerge naturally when the rear axle loses traction and when an oncoming vehicle requires aggressive evasion.
    \item Through comparisons with conventional stability control, we highlight a tradeoff between open-loop stability and steering authority, and demonstrate via a Monte Carlo study that drift-capable control achieves better lane keeping at a range of speeds, though with a small risk of more severe road departures on long patches of ice.
\end{itemize}
\noindent To the best of our knowledge, this is the first study of emergent autonomous drifting behavior in scenarios derived from real-world crash data.  Ultimately, this work provides evidence that drifting can be an optimal strategy for collision avoidance in realistic winter driving scenarios.

\section{Vehicle Dynamics Model}
\label{sec:dynamics}

To capture dynamics in winter conditions, we use a nonlinear single-track vehicle model suitable for highly dynamic driving on low and variable friction surfaces \cite{weber2024human, lew2025risk} with the following states: yaw rate $r$, sideslip angle $\beta$, total velocity at the center of mass $V$, longitudinal weight transfer $\Delta F_z$, front wheelspeed $\omega_f$, rear wheelspeed $\omega_r$, longitudinal path position $s$, heading error of velocity vector $\Delta \phi$, and lateral path position $e$. Modeled as an all-wheel drive (AWD) vehicle with independent front and rear brakes, the control inputs are: steering angle $\delta$, total engine torque $\tau_\text{eng} \geq 0$ (decomposed into front $\tau_{\text{eng},f} = \tau_\text{eng} f_\text{eng}$ and rear $\tau_{\text{eng},r} = \tau_\text{eng} (1-f_\text{eng})$ torques according to constant engine bias $f_\text{eng} = 0.4$), front brake torque $\tau_{\text{brake},f} \leq 0$, and rear brake torque $\tau_{\text{brake},r} \leq 0$.

Using physics-based modeling of vehicle motion along a curved path, the following system of coupled differential equations describes the evolution of vehicle states:

\begin{equation}
\begin{aligned}
\dot{r} &= \tfrac{1}{I_z}\left[aF_{xf}\sin\delta + aF_{yf}\cos\delta - bF_{yr}\right] \\
\dot{\beta} &= \tfrac{1}{mV}\left[F_{xf}\sin(\delta{-}\beta) + F_{yf}\cos(\delta{-}\beta) - F_{xr}\sin\beta\right. \\
            &\phantom{{}=\tfrac{1}{mV}\left[\right.}+ \left.F_{yr}\cos\beta - F_\text{drag}\sin\beta\right] - r \\
\dot{V} &= \tfrac{1}{m}\left[F_{xf}\cos(\delta{-}\beta) - F_{yf}\sin(\delta{-}\beta) + F_{xr}\cos\beta\right. \\
         &\phantom{{}=\tfrac{1}{m}\left[\right.}+ \left.F_{yr}\sin\beta + F_\text{drag}\cos\beta\right] \\
\Delta\dot{F}_z &= -K\left[\Delta F_z - \tfrac{h_\text{cg}}{a+b}\left(F_{xr} + F_{xf}\cos\delta - F_{yf}\sin\delta\right)\right] \\
\dot{\omega}_f &= \tfrac{1}{I_{wf}}\left[\tau_{\text{eng},f} + \tau_{\text{brake},f} - F_{xf}R_f\right] \\
\dot{\omega}_r &= \tfrac{1}{I_{wr}}\left[\tau_{\text{eng},r} + \tau_{\text{brake},r} - F_{xr}R_r\right] \\
\dot{s} &= \tfrac{V\cos(\Delta\phi)}{1 - \kappa e}, \quad
\Delta\dot{\phi} = \dot{\beta} + r - \kappa\dot{s}, \quad
\dot{e} = V\sin(\Delta\phi),
\end{aligned}
\label{eq_state_dyns}
\end{equation}


\noindent where $a$ and $b$ are the distances from the front and rear axles to the center of mass, $I_z$ is the vehicle's z-axis moment of inertia, $m$ is the vehicle's mass, $F_\text{drag}$ is a velocity-dependent drag force, $K$ represents the vehicle's longitudinal suspension stiffness, $h_\text{cg}$ is the height of the vehicle's center of mass, $I_{wf}$ and $I_{wr}$ are the front and rear axle moments of inertia, $R_f$ and $R_r$ are the wheel radii, and $\kappa$ is the local path curvature.

To accurately capture the effects of engine and brake torques in an AWD vehicle model up to the friction limits, we compute the tire forces on both axles $F_{xf}$, $F_{yf}$, $F_{xr}$, and $F_{yr}$ with a coupled slip brush tire model \cite{pacejka2005tire}. In the coupled slip model, the total tire force on each axle is:
\begin{equation}
F = 
\begin{cases}
\sigma C -\tfrac{\sigma^2C^2}{3\mu F_z}+\tfrac{\sigma^3C^3}{27\mu^2F_z^2},& \sigma \leq \tan^{-1} \left( \tfrac{3\mu F_z}{C} \right) \\
\mu F_z,& \sigma > \tan^{-1} \left( \tfrac{3\mu F_z}{C} \right),
\end{cases}
\label{eq_total_tire_force}
\end{equation}

\noindent where $\sigma = \sqrt{\lambda^2+\tan{(\alpha)}^2}$ is the total slip from the combination of longitudinal slip $\lambda$ and lateral slip angle $\alpha$. This tire model captures the inherent coupling of longitudinal and lateral forces in drifting maneuvers as the tires transition from linear deformation defined by isotropic tire stiffness $C$ to saturation at the total available friction force $\mu F_z$, where $\mu$ is the friction coefficient, and $F_z$ is the normal force on each axle computed using weight transfer state $\Delta F_z$. This total tire force is broken into lateral and longitudinal components:
\begin{equation}
    F_x = \tfrac{\lambda}{\sigma}F,
    \quad 
    F_y = \tfrac{-\tan{(\alpha)}}{\sigma}F,
    \label{eq_tire_Fx_Fy}
\end{equation}
where the slips on each axle are defined in terms of vehicle states and inputs as:
\begin{equation}
\begin{aligned}
\lambda_f =& \tfrac{\omega_f R_f - [V \cos{(\beta)} \cos{(\delta)} + (V \sin{(\beta)} + ar)\sin{(\delta)}]}{|V \cos{(\beta)} \cos{(\delta)} + (V \sin{(\beta)} + ar)\sin{(\delta)}|} \\
\lambda_r =& \tfrac{\omega_r R_r - V \cos{(\beta)}}{|V \cos{(\beta)}|} \\
\alpha_f =& \tan^{-1} {\left(\tfrac{V \sin{(\beta)} + ar}{V \cos{(\beta)}}\right)} - \delta \\
\alpha_r =& \tan^{-1} {\left(\tfrac{V \sin{(\beta)} - br}{V \cos{(\beta)}}\right)}.
\end{aligned}
\label{eq_tire_slips}
\end{equation}

\section{Drift-Capable MPC Controller}
\label{sec:controller}

We seek an intuitive controller for winter driving that can safely operate at the friction limits on both axles and preemptively avoid hazards ahead on the road, including other moving vehicles. We therefore present a single-horizon nonlinear MPC framework with a simple cost structure that incentivizes safe winter driving, taking inspiration from the compact cost function in Weber \textit{et al.}’s winter rally controller \cite{weber2024human}. This allows the vehicle to enter high sideslip drifting states when needed, without explicitly tracking drifting equilibria.

The controller repeatedly computes a trajectory of vehicle states $\mathbf{x}^i = [r^i, \beta^i, V^i, \Delta F_z^i, \omega_f^i, \omega_r^i, s^i, \Delta \phi^i, e^i]^\top$ and control inputs $\mathbf{u}^i = [\delta^i, \tau_\text{eng}^i, \tau_{\text{brake},f}^i, \tau_{\text{brake},r}^i]^\top$ over a finite time horizon by solving the following optimization problem:

\begin{alignat}{2}
\min_{\mathbf{x}, \mathbf{u}} \quad & \sum_{i=0}^{N} \Big(J^i_\text{path} + J^i_\text{speed} + J^i_\text{heading} + J^i_\text{input} + J^i_\text{safety}\Big) \notag \\
\text{s.t.} \quad & \mathbf{x}^{i+1} = \mathbf{x}^i + \tfrac{\Delta t}{2}[\dot{\mathbf{x}}^i + \dot{\mathbf{x}}^{i+1}], && \hspace{-2cm} i = 0,\dots,N-1, \notag \\
& (\mathbf{x}^0,\, \mathbf{u}^0) = (\mathbf{x}_{\text{meas}},\, \mathbf{u}_{\text{meas}}), \notag \\
& \mathbf{u}_{\min} \le \mathbf{u}^i \le \mathbf{u}_{\max}, && \hspace{-2cm} i = 1,\dots,N, \notag \\
& \omega_{f,r}^i \le \omega_{\max}, && \hspace{-2cm} i = 1,\dots,N. \label{eq_opt_problem}
\end{alignat}

\noindent In the cost function, the first three terms encode the objectives of driving in the center of the right lane $J^i_\text{path} = W_e (e^i-e_\text{ref})^2$, tracking a desired speed $J^i_\text{speed} = W_V (V^i-V_\text{ref})^2$, and keeping the velocity vector pointed along the path $J^i_\text{heading} = W_{\Delta \phi} (\Delta \phi^i)^2$, with quadratic penalties weighted by coefficients $W_j$. The control input cost term $J^i_\text{input} = (\mathbf{u} ^{i}) ^\top R \mathbf{u}^i + (\dot{\mathbf{u}} ^{i}) ^\top S \dot{\mathbf{u}}^i$ lightly penalizes large control inputs and input rates via quadratic costs with diagonal matrices $R$ and $S$. This cost helps prevent simultaneous braking and throttle on the same axle unless needed to satisfy other objectives. Lastly, the safety term $J^i_\text{safety}$ places a high cost on the vehicle if it goes outside the road edges or collides with an oncoming vehicle, with collisions penalized more heavily than road departures to reflect their greater severity. In the safety cost, we treat the vehicle as two circles and apply a quadratic penalty on penetration of these two circles with the road edges and with the oncoming vehicle as in \cite[Eqs. (12), (18), (19)]{brown2019coordinating}.

The first constraint embeds the nonlinear dynamics of the vehicle into the MPC problem via an implicit trapezoidal discretization of the continuous system in (\ref{eq_state_dyns}) using a timestep $\Delta t = 0.12$ s. Over an $N = 32$ stage horizon, the vehicle plans $96$ m ahead on the road when driving at $55$ mph ($25$ m/s). The remaining constraints ensure that the MPC horizon starts at the measured states and control inputs for each solve, and that control inputs and wheelspeeds stay within limits imposed by the vehicle's available actuation. We emphasize that there are no costs or constraints on sideslip or yaw rate states. The vehicle is free to drift as needed to achieve the objectives in the cost function but is never explicitly rewarded to do so.

The problem (\ref{eq_opt_problem}) is solved online with costs and constraints using a standard sequential quadratic programming (SQP) scheme \cite[Chapter 18]{Nocedal2006}, where the dynamics and gradients are computed using Python's \texttt{Jax} library \cite{bradbury2018jax}, and the inner quadratic programs are solved with \texttt{OSQP} \cite{stellato2020osqp}. This approach enables repeated replanning with sub-$20$ ms average solve times for real-time operation on a laptop with an Intel Core i9-13900H CPU.

\section{Real-World Scenarios}
\label{sec:scenario}

To examine whether drifting behavior naturally emerges from our MPC controller in real-world winter scenarios, we extract the most common scenarios that lead to fatal crashes in winter conditions on United States (US) public roads and implement these scenarios with realistic parameter values. 

\subsection{NHTSA Crash Clusters}

The US National Highway Traffic Safety Administration (NHTSA) compiles annual statistics on road vehicle fatalities and non-fatal crashes based on police reports. This aggregated set of crashes provides the best available data on overall US crash trends across a variety of scenario-relevant variables. Using this data from NHTSA, Weiss \textit{et al.} filter fatal crashes between $2011$-$2022$ for those involving light vehicles in winter conditions and create representative scenario clusters \cite{weiss2025mixed}. Their results show that the majority of fatal winter crashes occur on flat $2$-lane, non-National Highway System (NHS), local-rural roads with $55$ mph speed limits. Approximately $50\%$ of fatal crashes involve a single vehicle losing control and departing the road, and a majority of multi-vehicle crashes involve a collision with an oncoming vehicle. Within the single-vehicle crashes, the two most common causes from police reports are excessive speed and suddenly encountering poor road conditions. Further, the dominant contributing factor to collisions with oncoming vehicles is that one of the vehicles loses control on a slippery road surface.

Based on these representative crashes from NHTSA data, we test the drift-capable controller in the following scenarios:
\begin{enumerate}\renewcommand{\theenumi}{S\arabic{enumi}}
\renewcommand{\labelenumi}{\textbf{\theenumi:}}
    \item A single-vehicle road departure on a snowy $2$-lane, $55$ mph road due to the vehicle initially traveling too fast for road conditions.
    \item A single-vehicle road departure on a snowy $2$-lane, $55$ mph road due to the vehicle suddenly encountering a patch of ice or other low friction section.
    \item A multi-vehicle collision with an oncoming vehicle on a snowy $2$-lane, $55$ mph road due to one of the vehicles losing control and sliding into the left lane.
\end{enumerate}
While the NHTSA data provides high-level, primarily qualitative crash descriptions, we need to parameterize these scenarios for practical implementation in simulations.

\subsection{Road Geometry}
The curvature and lane width of the road are critical parameters that contribute to the risk of a crash. To capture the geometry of real roadways for our winter scenarios, we look to road design manuals published by Departments of Transportation (DOT) in states that experience significant snowfall. Based on the $1991$-$2020$ US National Oceanic and Atmospheric Administration (NOAA) climate normals, $3$ of the $5$ snowiest weather stations representing populations of at least $30{,}000$ are in New York \cite{noaa_normal_snowfall_1991_2020}.

\begin{figure}[t]
  \centering
  \includegraphics[width=\linewidth, trim=1.5cm 1cm 5.5cm 4cm, clip]{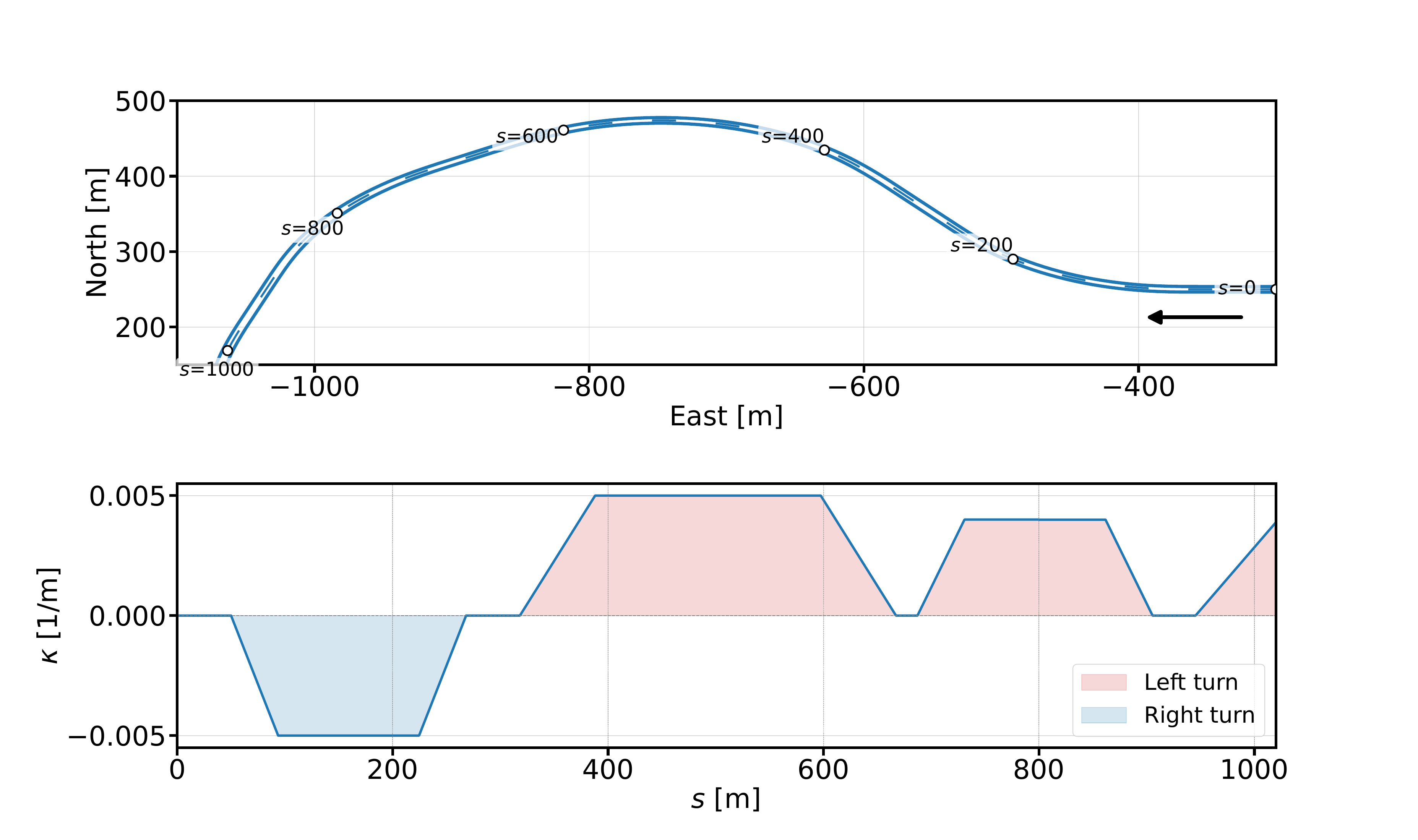}
  \caption{Position coordinates (top) and curvature (bottom) of the $2$-lane road based on NY state road design guidance.}
  \label{fig_roadway}
\end{figure}

For non-NHS local-rural roads with a design speed of $55$ mph, the New York (NY) Highway Design Criteria specify a minimum radius of curvature of $651$ ft ($198$ m) and a $12$ ft ($3.7$ m) travel lane width for roads where ``crash rate is above the statewide rate" \cite{nydot_hdm_ch2_design_criteria_2025}. Further qualitative guidance provided by the NY Highway Basic Design manual suggests the use of spiral curves of ``constantly changing radius" to provide a smooth path for vehicles entering and exiting circular curves and to reduce the need for lane widening along horizontal curves \cite{nydot_hdm_ch5_basic_design_2025}. The $2$-lane road in our scenarios, which is designed to meet minimum radius, lane width, and constant radius transition guidance, is shown in spatial coordinates and as a curvature profile over longitudinal position $s$ in Fig. \ref{fig_roadway}.

\subsection{Friction Coefficient}
The available friction between the tires and the road $\mu$ is another critical parameter for studying the behavior of vehicles in real-world winter scenarios. Weiss \textit{et al.} derived a distribution of friction coefficients from winter control loss events recorded by instrumented vehicles in the Second Strategic Highway Research Program (SHRP2) naturalistic driving study \cite{weiss2025mixed}. This distribution provides an approximation for baseline friction values experienced by vehicles when losing traction on real roadways. We take the mean value of approximately $0.4$ from this naturalistic distribution and use it as a constant friction coefficient representing snowy conditions across all scenarios. Notably, this value falls within a range of friction coefficients measured or estimated with real vehicles on snowy roads in other winter driving studies \cite{pu2021road, ding2022winter, volkmann2024friction}.

In addition to a lower baseline friction coefficient, hidden patches of ice and other sudden friction variations (e.g. due to slush) represent clear dangers in winter weather; rapidly changing road surface conditions are one of the main causes of winter fatalities tabulated in NHTSA crash data. Several researchers have recorded local friction variations in winter conditions with onboard estimators \cite{erdogan2009adaptive} and optical road surface sensors \cite{maanpaa2024dense}, showing that friction coefficients can suddenly decrease to $0.1$-$0.2$ on patches of ice. We choose to model ice patches as drops in friction with $\mu = 0.1$.

\section{MPC Simulation Results}
\label{sec:results}

Due to the inherent danger of testing high speed, multi-vehicle winter scenarios in real vehicles and the limited availability of winter testing facilities with sufficient space for $55$ mph experiments, we test the MPC controller in a high fidelity simulator.\footnote{A video of the simulation results in a snowy virtual environment is available here: \url{https://youtu.be/LKihrnLmTgM}.} Our simulator dynamics are a double-track extension of the AWD model in (\ref{eq_state_dyns}), which matches the fidelity of reference models used to emulate low friction surfaces in various experimental studies \cite{weiss2022combining, dallas2025control}. The vehicle parameters in simulation approximate those of a Lexus LC 500 with AWD for handling low friction winter conditions.

\subsection{Driving Through Curved Sections at a Safe Speed}
In scenario \textbf{S1}, we investigate the controller's ability to slow down the vehicle to a safe speed for the road conditions. We assume that the vehicle has knowledge of roadway curvature over the MPC prediction horizon and of the baseline friction value $\mu = 0.4$. We set the reference speed $V_\text{ref}$ to $32$ m/s, even though the vehicle needs to slow down below $28$ m/s to stay in the lane (using a point mass approximation, the maximum speed possible for lane keeping on the sharpest curve within the vehicle's friction limits is $\sqrt{\mu g R_\text{min}} = \sqrt{0.4 \times 9.81 \text{m}/\text{s}^2 \times 200 \text{m}} = 28.0$ m/s).

\begin{figure}[t]
  \centering
  \includegraphics[width=\linewidth, trim=2cm 0.5cm 4.5cm 4cm, clip]{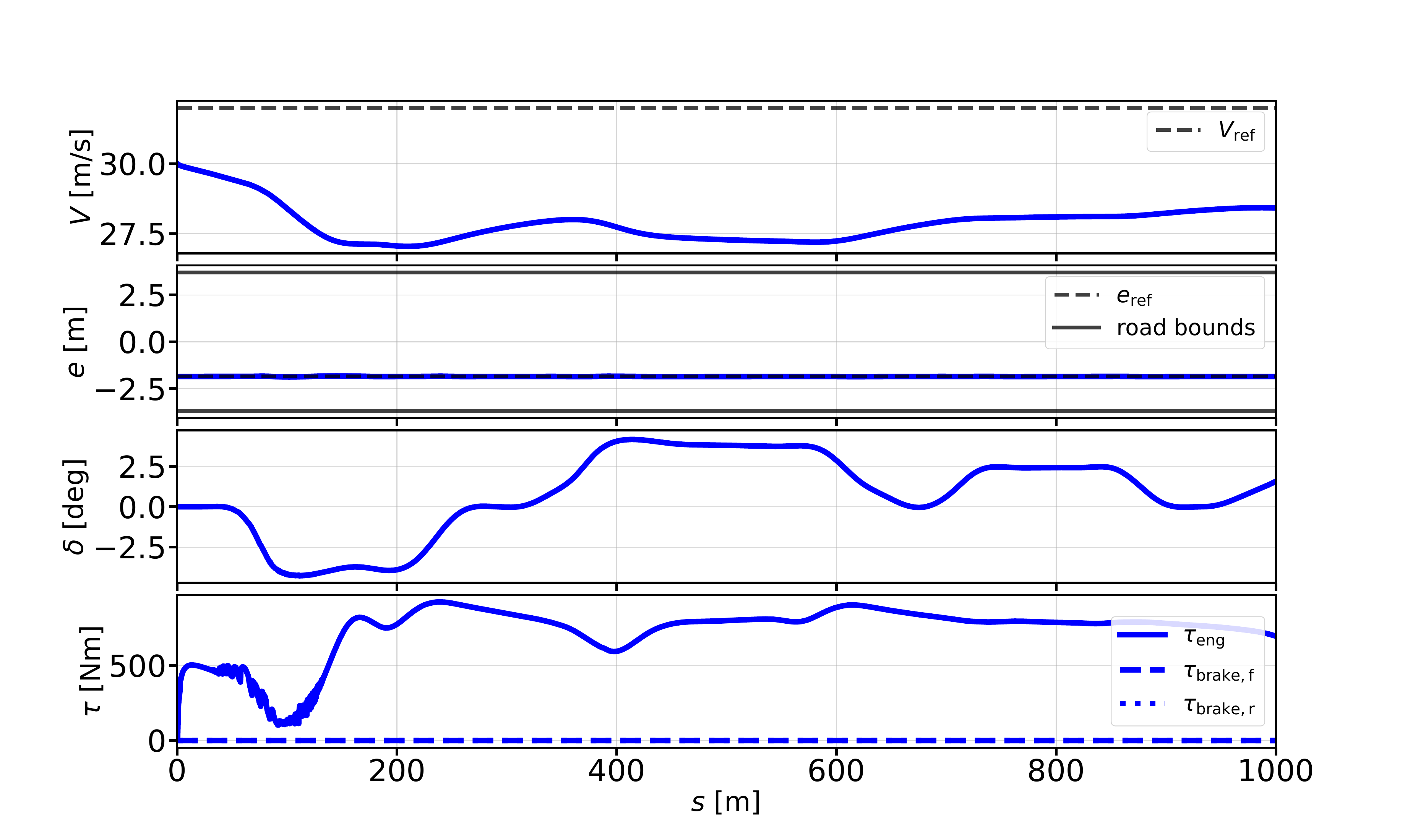}
  \caption{States and controls for the MPC controller while slowing down for curved sections of the road in scenario \textbf{S1}.}
  \label{fig_slowdown_states_ctls}
\end{figure}

The plots in Fig. \ref{fig_slowdown_states_ctls} show the vehicle speed, path position, steering angle, and engine/brake torques, demonstrating how the controller slows the vehicle down to a safe speed to ensure path tracking around the turns. Since the MPC controller can look ahead into the turn, it can preemptively reduce velocity approaching curved sections of the road. In this scenario with a constant friction surface and large radius of curvature, the vehicle stays safe by simply slowing down. 

\subsection{Drifting-Aided Recovery from Ice Patches}
\label{subsec:ice}

In scenario \textbf{S2}, we study the vehicle's ability to handle unexpected patches of ice. In this simulation, the vehicle encounters short patches of ice with $\mu = 0.1$ on the front axle, the rear axle, and both axles, while the MPC controller's assumed friction coefficient remains unchanged at $\mu = 0.4$. We set the desired speed to $25$ m/s, matching the $55$ mph speed limit. Although this speed is slow enough to handle the curved road sections under nominal friction conditions, it is too high for the ice patches. 

\begin{figure}[t]
  \centering
  \vspace*{-1.5pt}
  \includegraphics[width=\linewidth]{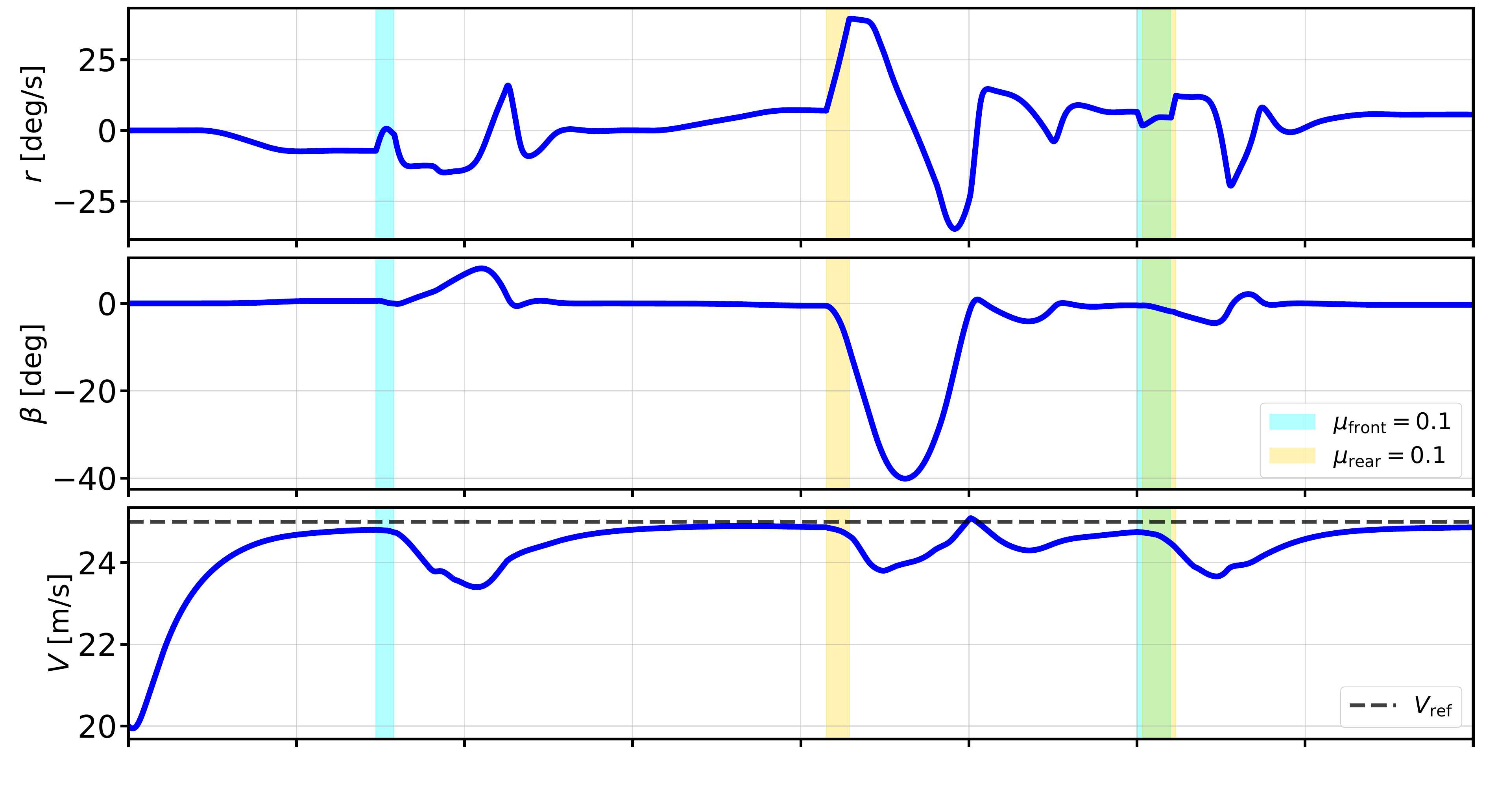}\\[-10.5pt]
  \begin{tikzpicture}
    \node[anchor=south west, inner sep=0] (main) at (0,0) 
      {\includegraphics[width=\linewidth]{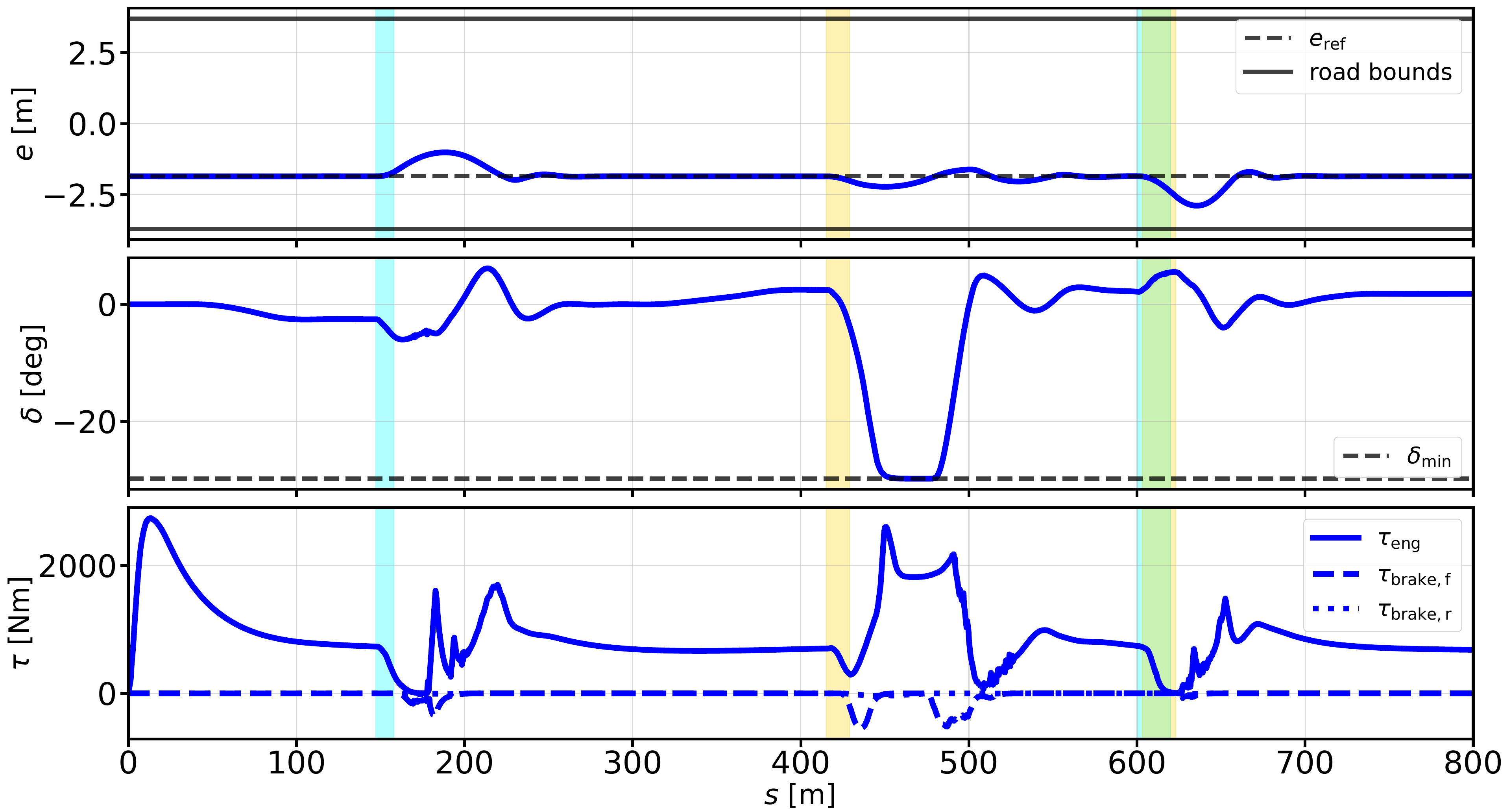}};
    \begin{scope}[x={(main.south east)}, y={(main.north west)}]
    \draw[gray, line width=1pt, draw opacity=0.7] (0.54, 0.098) rectangle (0.65, 0.15);
    \node[anchor=south west, inner sep=0, draw=gray, line width=2pt, draw opacity=0.7] (inset) 
        at (0.28, 0.3) 
        {\includegraphics[trim={32pt 32pt 115pt 80pt}, clip, width=0.25\linewidth]
          {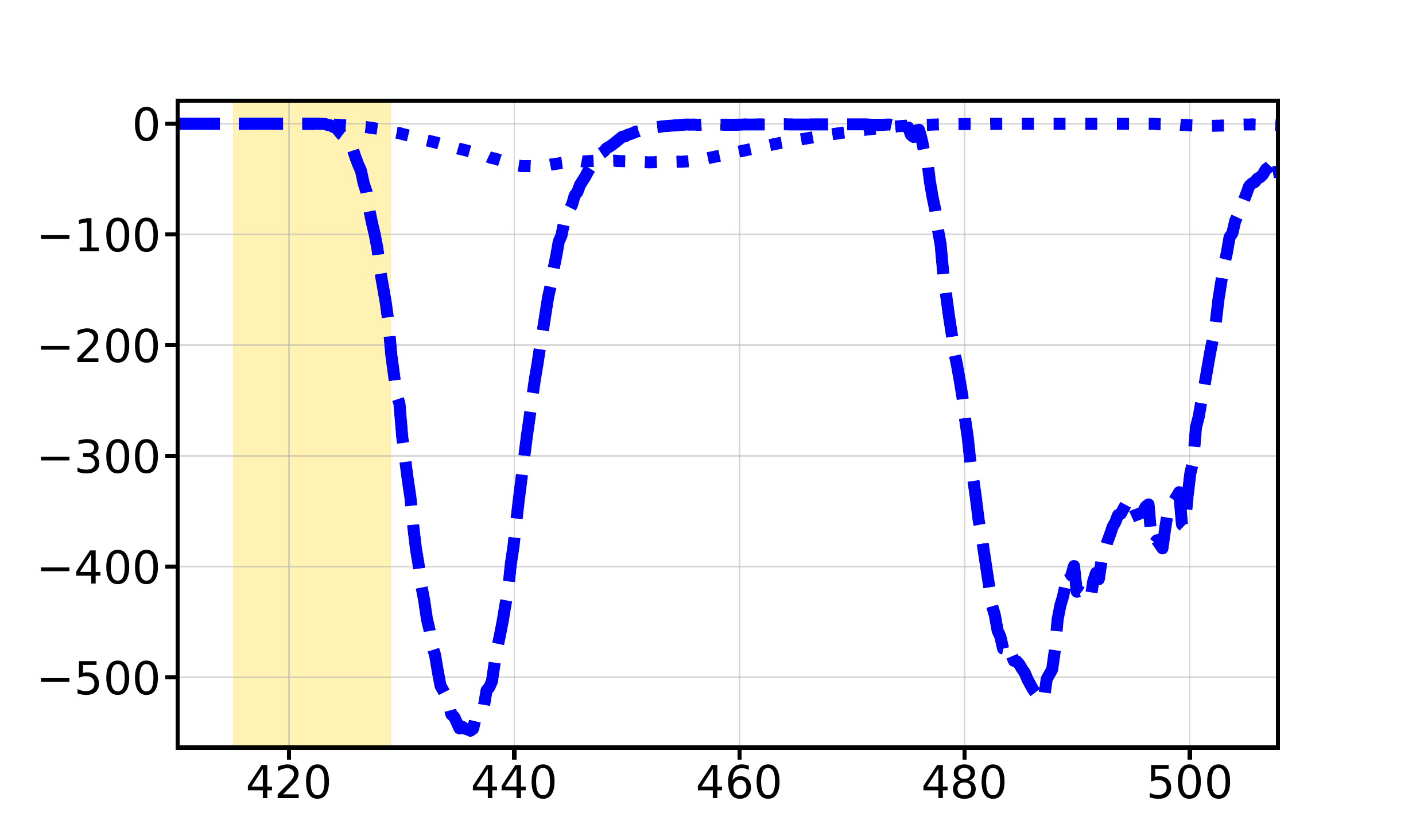}};
      \draw[gray, line width=1pt, draw opacity=0.7] (0.65, 0.15) -- (inset.north east);
      \draw[gray, line width=1pt, draw opacity=0.7] (0.54, 0.15) -- (inset.south west);
    \end{scope}
  \end{tikzpicture}
  \caption{MPC states and controls navigating scenario \textbf{S2} with ice patches, showing an emergent drifting maneuver up to $|\beta|_{\max} = 40^\circ$ between $s = 415$-$500$ m. The green background indicates ice patches on both axles.}
  \label{fig_icepatch_states_ctls}
\end{figure}
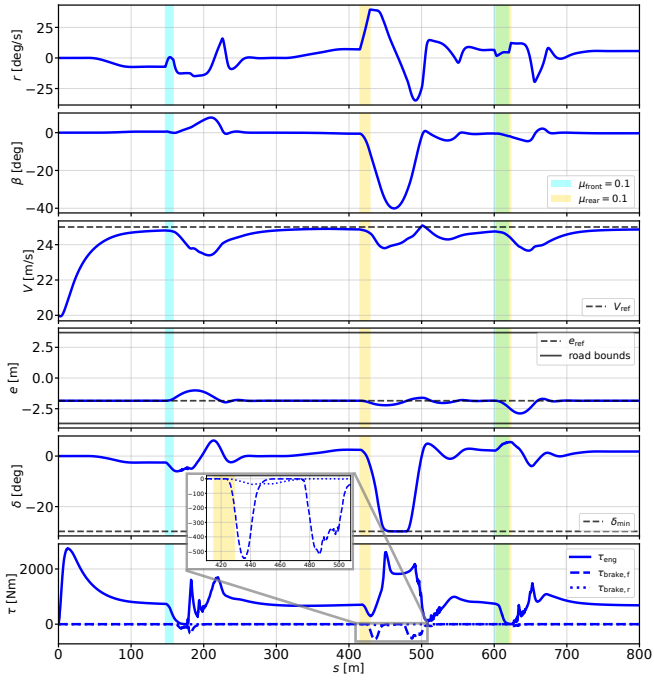

Fig. \ref{fig_icepatch_states_ctls} shows plots of the states and control inputs throughout this scenario. When the front axle is briefly on the ice patch at $s = 147$ m, the vehicle understeers to the left, which the MPC controller counteracts with a slight steering correction and brake application. These inputs slow down the vehicle's slide to reduce its excursion away from the desired path. The vehicle behaves similarly when both axles hit ice patches at $s = 600$ m, in this case sliding to the right since the road is curving to the left at that location.


While these instances of front axle tire force saturation result in simple corrections, the controller executes a much more complex, high sideslip drifting maneuver in response to the rear axle driving over a patch of ice at $s = 415$ m. Within a very short period of time, the vehicle momentarily engages the brakes to slow the vehicle down and transfer weight forward ($s = 435$ m), applies significant throttle and full countersteer ($s = 450$-$490$ m), and briefly brakes on the front axle while straightening out from the drift to prevent overcorrecting in its rotation back to the right ($s = 485$ m). Remarkably, throughout this emergent drifting maneuver, the vehicle exceeds $40^{\circ}$ of sideslip angle at close to $55$ mph while only deviating from the lane center by $37$ cm. With the simple objective of keeping the vehicle moving along the right lane and avoiding off-road excursions, the nonlinear MPC controller executes a highly skillful drifting maneuver. 

\subsection{Avoiding Severe Crashes with an Oncoming Vehicle}
We lastly show the drift-capable controller avoiding an oncoming vehicle in scenario \textbf{S3}. The MPC controller has knowledge of the other vehicle's current velocity and position states and assumes constant linear and rotational accelerations for the oncoming vehicle over the MPC prediction horizon. In this scenario, the other vehicle drives in its right lane toward the ego vehicle (running MPC) at $25$ m/s and then hits an unexpected patch of ice on both axles as the vehicles are approaching each other on a curved section of the road.

\begin{figure}[t]
  \centering
  \includegraphics[width=\linewidth, trim=0.5cm 0.5cm 0.3cm 0.5cm, clip]{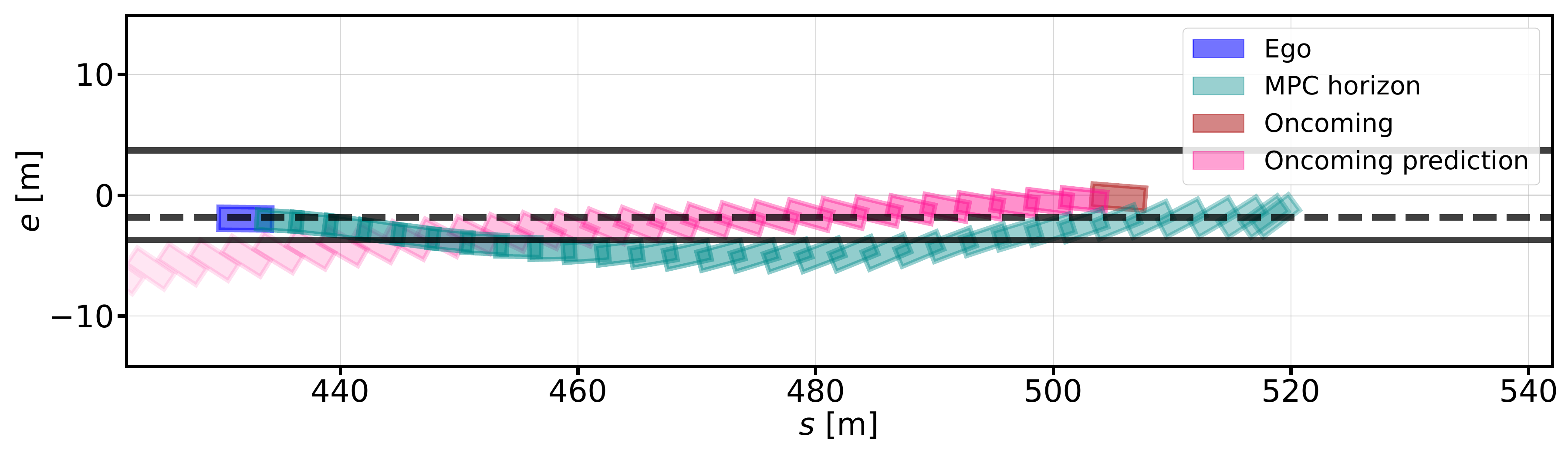}\\[-1.2em]
  \includegraphics[width=\linewidth, trim=0.5cm 0.5cm 0.3cm 0.5cm, clip]{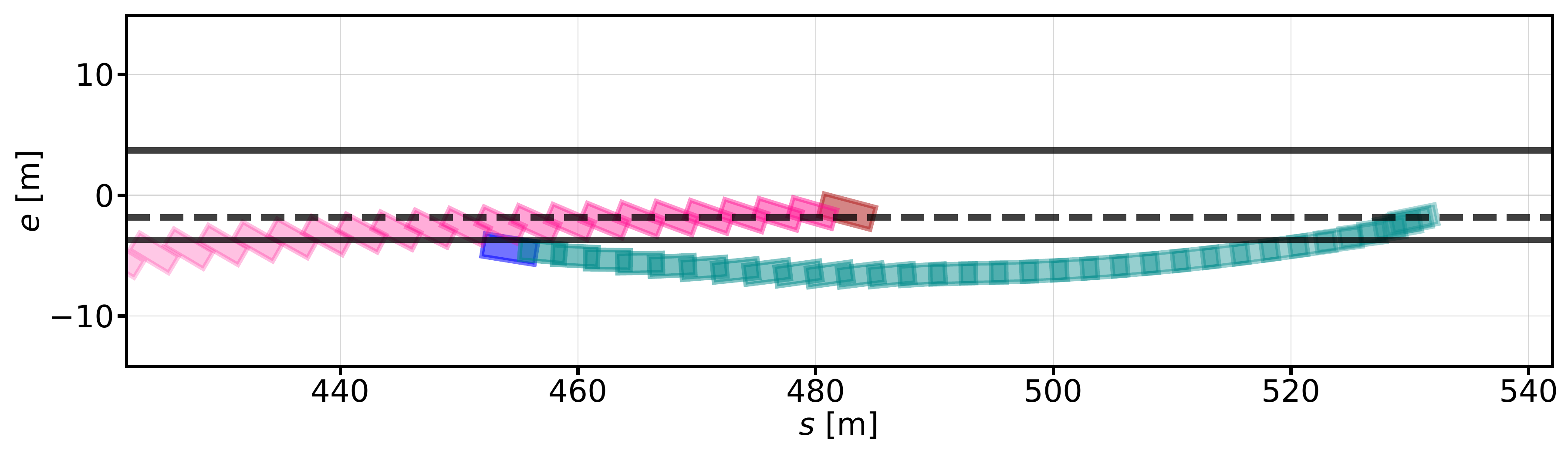}\\[-1.2em]
  \includegraphics[width=\linewidth, trim=0.5cm 0cm 0.3cm 0.5cm, clip]{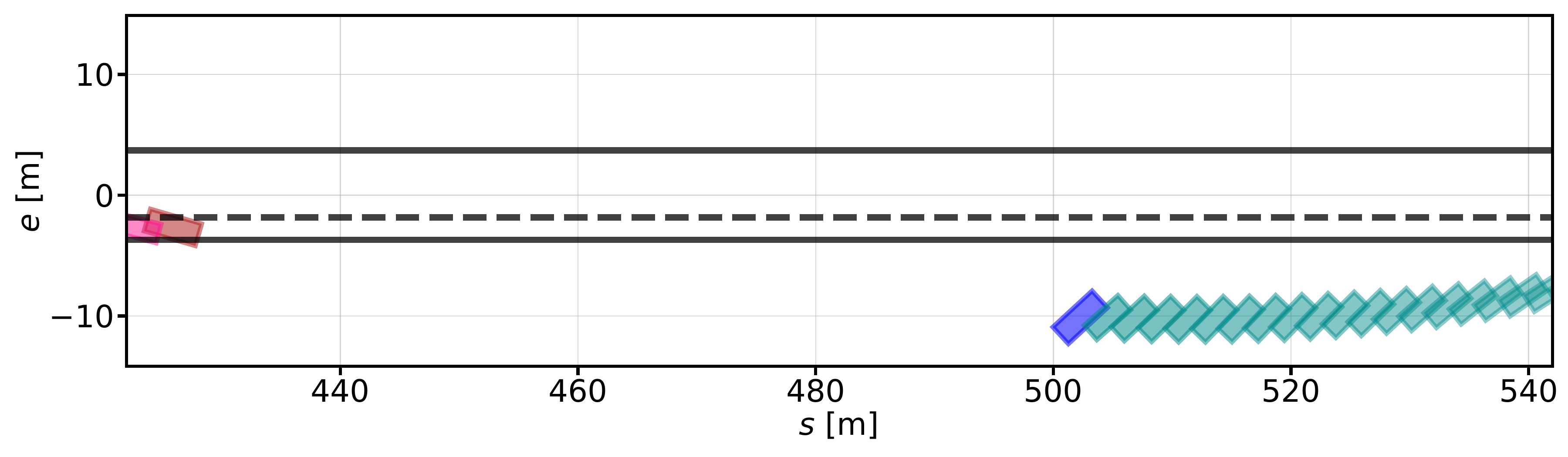}
  \caption{MPC snapshots when avoiding an oncoming vehicle that slides directly into the ego vehicle's lane in scenario \textbf{S3}. The ego vehicle returns to the road with a drifting maneuver reaching $|\beta|_\text{max} = 53^\circ$ (bottom).}
  \label{fig_avoidado_svse}
\end{figure}

Slipping to the left due to contact with the ice patch, the oncoming vehicle attempts to maneuver back to its lane, countersteering across the ego vehicle's path. This situation puts the MPC controller in a difficult position, ultimately requiring a drastic drifting maneuver to prevent a crash, as shown in the spatial coordinate plots in Fig. \ref{fig_avoidado_svse}. Choosing to avoid the much higher cost of an oncoming vehicle collision, the controller throws the vehicle off the road with a quick steering input to the right. The vehicle then slides sideways, which the controller sustains by applying throttle and countersteering, reaching a maximum sideslip angle of $53^{\circ}$ (Fig. \ref{fig_avoidado_svse}, bottom). Although the vehicle travels nearly $6.3$ m off the road, it is able to quickly return to the lane through this emergent drifting maneuver and continue driving after avoiding the oncoming vehicle.

\section{Comparison with Stability Control}
\label{sec:comparison}

To contextualize the drift-capable MPC controller with comparisons to a more conventional stability control system found on production vehicles, we implement a baseline electronic stability control (ESC) system as a safety filter on a feedforward-feedback autonomous lane keeping and cruise control (LK-CC) system. The ESC system determines the difference between the intended yaw rate from the LK-CC system (computed as $r_\text{intended} = \delta\tfrac{V \cos\beta}{a + b + K_\text{us}(V\cos\beta)^2}$, where $K_\text{us}$ is the vehicle's understeer gradient) and the measured yaw rate. When this error exceeds a threshold of $0.1$ rad/s, the system intervenes by cutting engine torque and applying brake torque to the outer front wheel to correct oversteer and the inner rear wheel to correct understeer \cite{chatzikomis2014comparison}. 

\subsection{Stability-Controllability Tradeoff}
\begin{figure}[t]
  \centering
  \includegraphics[width=\linewidth, trim=2.5cm 1cm 5cm 4.2cm, clip]{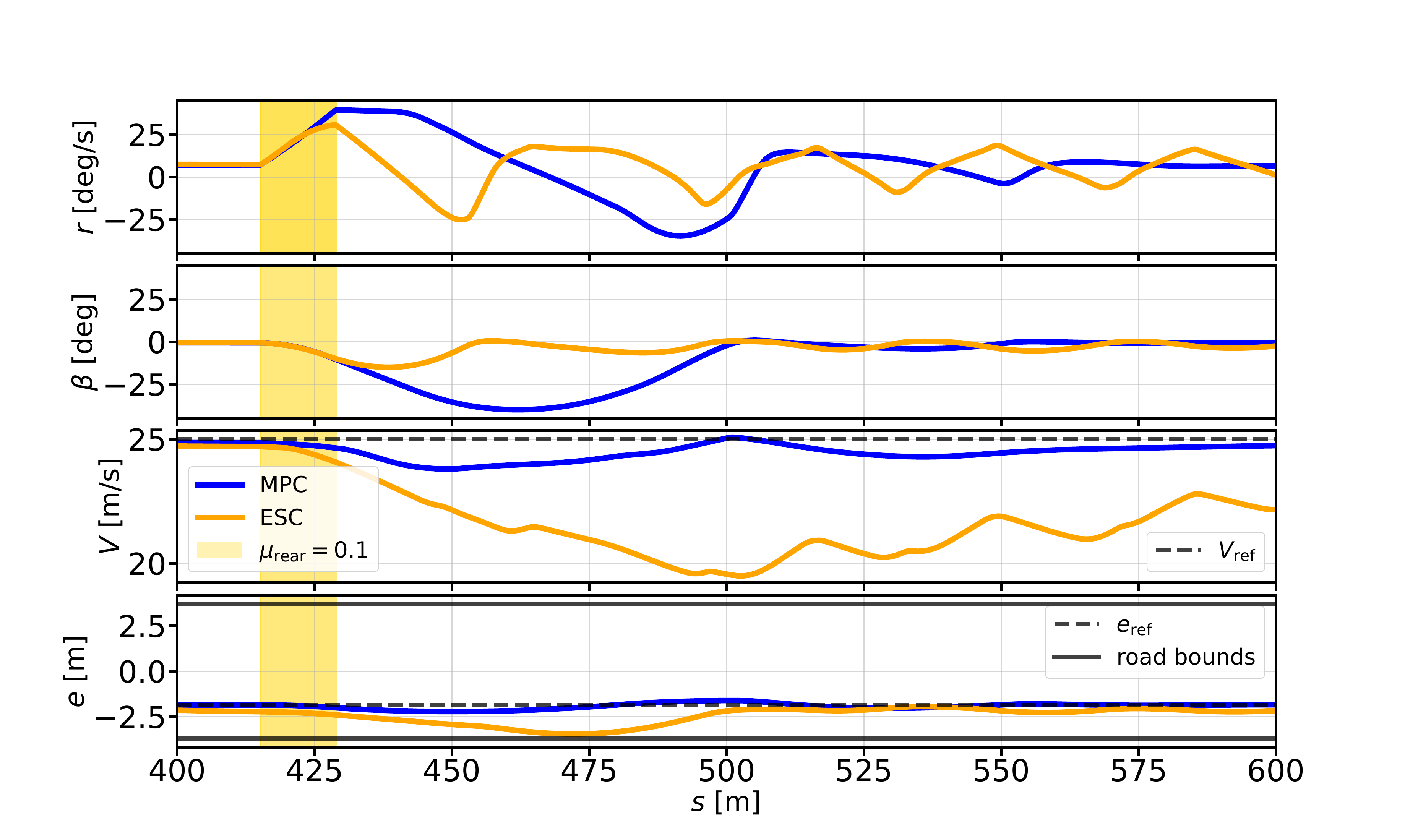}
  \caption{Comparing MPC (blue) vs ESC (orange) vehicle states after hitting a rear axle ice patch in scenario \textbf{S2}.}
  \label{fig_icepatch_compare_states}
\end{figure}

Our first comparison analyzes both controllers after the rear ice patch in scenario \textbf{S2}. While the drift-capable MPC controller executes a high sideslip maneuver to reduce lateral path error, the ESC system's top priority is to reduce oversteer by cutting throttle and providing corrective braking. The results in Fig. \ref{fig_icepatch_compare_states} show a comparison of the two controllers' behaviors after hitting a patch of ice on the rear axle. In quickly reducing the initial positive yaw rate and preventing significant sideslip, the ESC system slows the vehicle down much more than the MPC controller. Critically, the vehicle with ESC incurs $1.6$ m of lateral error ($4\times$ the lateral error of the MPC controller), nearly going off the edge of the road.

One way of understanding this difference in lane keeping ability between the two systems is to explore the stability-controllability tradeoff inherent to vehicle drifting \cite{hindiyeh2014controller, karino2023shared}. To do so, we linearize the closed-loop system for each controller with a numerical first-order Taylor series expansion around the states at each timestep $t$, resulting in a system of the form $\dot{\mathbf{x}}^t = A^t \mathbf{x}^t + B^t \mathbf{u}^t$. The eigenvalues of the $A$ matrix determine local stability; if any eigenvalue has a positive real part, the system is locally unstable. The magnitudes of the entries of the $B$ matrix indicate how much control authority each input has over each state. 

\begin{figure}[t]
  \centering
  \includegraphics[width=\linewidth, trim=2.5cm 1cm 5cm 4.2cm, clip]{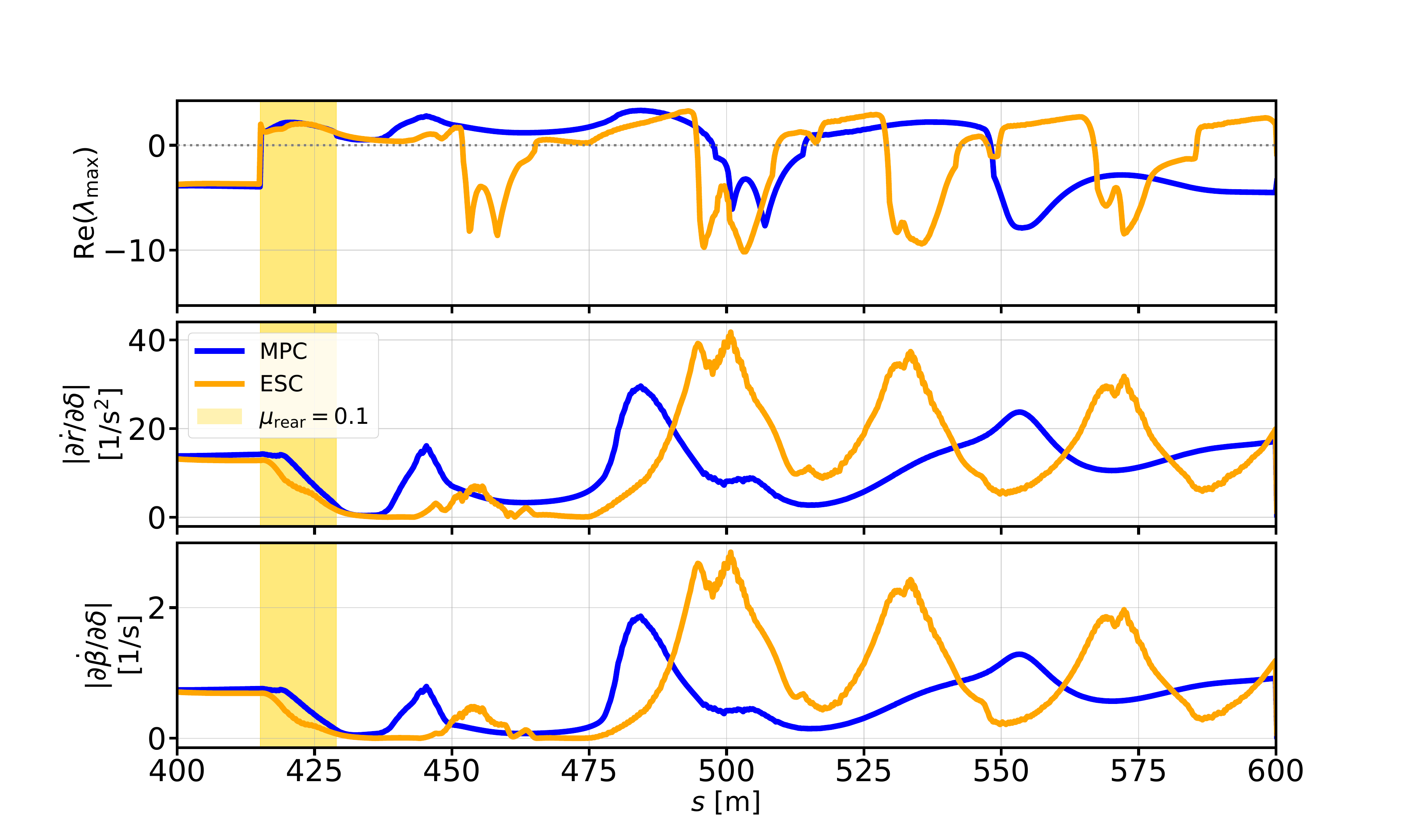}
  \caption{Metrics of local stability and steering control authority for MPC (blue) vs ESC (orange) controllers after the rear ice patch in scenario \textbf{S2}.}
  \label{fig_icepatch_compare_stab_ctl}
\end{figure}

The plots in Fig. \ref{fig_icepatch_compare_stab_ctl} show the real part of the largest eigenvalue of the $A$ matrices and the partial derivatives of yaw acceleration $\dot{r}$ and sideslip rate $\dot{\beta}$ with respect to steering input $\delta$. During the period of high sideslip drifting between $s = 415$-$500$ m, the MPC controller maintains a locally unstable open-loop system, while the ESC controller repeatedly drives the system into temporary local stability. This region of instability corresponds to greater steering control authority over lateral states $r$ and $\beta$ than ESC, enabling the MPC controller to maneuver the vehicle much closer to the desired position in the middle of the right lane, despite the ice patch disturbance. Once the ESC-controlled vehicle has sufficiently slowed down to be more easily directed back to the path, its steering control authority exceeds that of the MPC controller, beginning at $s = 490$ m. At this point, however, the vehicle has already veered significantly away from the center of the lane. The result in this example suggests that tolerating open-loop instability, rather than actively suppressing it, can provide the additional control authority needed to safely navigate roads with friction variability in winter conditions.

\subsection{Monte Carlo Study on Variable Friction Surfaces}
We further run simulations with a variety of randomly generated ice patches in scenario \textbf{S2} to study the robustness and overall safety of the drift-capable MPC controller compared with ESC. We create $100$ sets of random ice patches, each containing a front, rear, and combined low friction region with length sampled from a $5$-$15$ m uniform distribution. The MPC and ESC controllers are simulated in these $100$ unique friction environments with desired speeds $V_\text{ref}$ of $15$, $20$, and $25$ m/s. The simulation terminates when the vehicle traverses $800$ m of the roadway or when lateral lane position exceeds $20$ m.

\begin{table}[t]
\centering
\caption{Monte Carlo ice patch study: medians and counts for $100$ runs.}
\label{tab_mc_results}
\small
\setlength{\arrayrulewidth}{0.1pt}
\renewcommand{\arraystretch}{1.3}
\setlength{\tabcolsep}{4pt}
\begin{tabular}{|c|c|c|c|c|c|}
\hline
$V_\mathrm{ref}$ & Controller & $\overline{|e{-}e_\text{ref}|}$ & $|e{-}e_\text{ref}|_\text{max}$ & $N_{|\beta|_\text{max}>10^\circ}$ & $N_\text{depart}$ \\
\hline\hline
\multirow{2}{*}{$15$\,m/s} & MPC & $0.004$\,m & $0.126$\,m & $0$ & $0$ \\
 & ESC & $0.023$\,m & $0.162$\,m & $0$ & $0$ \\
\hline
\multirow{2}{*}{$20$\,m/s} & MPC & $0.013$\,m & $0.311$\,m & $13$ & $0$ \\
 & ESC & $0.076$\,m & $0.440$\,m & $3$ & $0$ \\
\hline
\multirow{2}{*}{$25$\,m/s} & MPC & $0.067$\,m & $0.847$\,m & $26$ & $3$ \\
 & ESC & $0.270$\,m & $1.170$\,m & $14$ & $7$ \\
\hline
\end{tabular}
\end{table}

Table \ref{tab_mc_results} presents results from the Monte Carlo study, showing the median values of mean and maximum lateral error from the center of the right lane $e_\text{ref}$. We also report the number of simulations (out of $100$) in which maximum sideslip exceeds $10^\circ$ and in which the vehicle drives outside the lane bound $|e{-}e_\text{ref}| > 1.85$ m. At lower speeds $V_\text{ref} = 15$ and $20$ m/s, the MPC controller drives with less lateral error by leveraging its knowledge of vehicle sideslip and prediction of upcoming road curvature, which the ESC system does not have, for more precise lane keeping. The MPC controller uses fewer large sideslip maneuvers to accomplish its objectives than at $V_\text{ref} = 25$ m/s, indicating that drifting maneuvers emerge as the optimal response to ice patches more often at higher speeds. For these simulations with lower reference speeds, both controllers manage the conditions safely; there are no road departures at $15$ and $20$ m/s.

\begin{figure}[t]
  \centering
  \includegraphics[width=\linewidth, trim=0.2cm 0.3cm 0.1cm 0.2cm, clip]{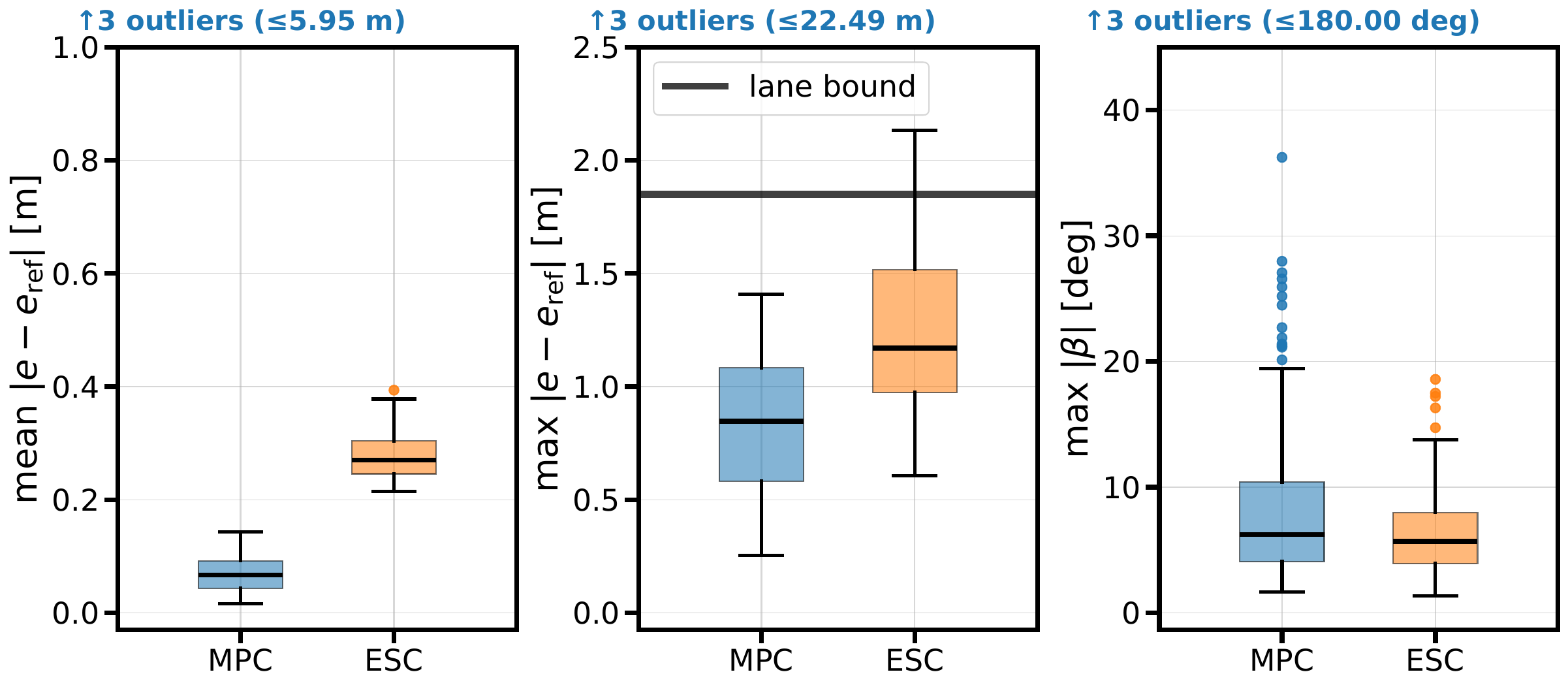}
  \caption{$V_\text{ref} = 25$ m/s Monte Carlo results showing mean and maximum lane error and maximum sideslip box plots, including $3$ MPC outliers.}
  \label{fig_mc_results}
\end{figure}

At $V_\text{ref} = 25$ m/s, both controllers experience lane departures, with MPC going out of the lane in $3$/$100$ simulations compared with $7$/$100$ for ESC. The nature of these road departures can be examined more closely in the box plots in Fig. \ref{fig_mc_results}, which show lateral error and sideslip distributions for the $25$ m/s condition. In the $3$ simulations with lane excursions, the MPC controller tries to use the full capabilities of the vehicle to recover from long rear ice patches ($> 14$ m in length), operating at peak countersteer and rear wheelspeed for too long to recover the drift, which ultimately results in lane departures of $15$-$20$ m. In the other $97$ simulations, the MPC controller operates with lower mean and maximum lane error while using higher sideslip maneuvers than ESC. Across all speeds, the drift-capable controller shows smaller lane keeping deviation and fewer out-of-lane events than the stability control system, but with the risk of occasional, significant road departures at higher speeds.

\subsection{Discussion: The Role of Speed in Safe Winter Driving}
In the Monte Carlo study, the MPC controller demonstrates a small risk of a serious road departure when tracking the highest speed $V_\text{ref} = 25$ m/s, while showing strong lane keeping performance without road departures at lower speeds. This finding raises the question of how conservatively an autonomous system should approach dangerous winter scenarios. Seeking more risk-averse control architectures, researchers have explored versions of MPC that plan with multiple models representing different possible friction coefficients in adverse weather conditions \cite{alsterda2019contingency, lew2025risk, aggarwal2025friction}. These controllers preemptively drive more carefully in anticipation of potential water puddles or ice patches, rather than reacting after the vehicle has gone over lower friction regions. Future autonomous systems may benefit from a combination of these approaches, proactively slowing down for dangers ahead, while permitting drifting maneuvers when needed to safely respond to extreme and unexpected disturbances.

\section{Limitations}
\label{sec:limits}
While this study presents a real-time MPC controller capable of high sideslip maneuvering in real-world winter scenarios, there are a few important limitations to note. First, the single-track vehicle model combines left and right tire forces on each axle, preventing the MPC controller from accounting for split-$\mu$ surfaces, which may occur in real winter conditions. This model also does not explicitly include lateral weight transfer, although the rally racing results in \cite{weber2024human} with a single-track model-based controller on a real winter testing surface suggest that roll motion dynamics are not required for precise control in high sideslip regimes. We further acknowledge the limitation of simulation-only validation and hope to run experiments on a full-size test vehicle in the future.

Additionally, this work assumes perfect sensing of the oncoming vehicle's motion, the nominal friction coefficient, and the ego vehicle's position and velocity states. These assumptions could be relaxed in future work, for example by implementing more realistic environmental sensor models and adding online friction estimation. As noted by Aggarwal and Gerdes, however, onboard estimators may not be able to suitably adjust the friction parameters of a motion planner within the time that the vehicle passes over the relatively short ice patches tested in this work \cite{aggarwal2025friction}. Further, simulating the oncoming vehicle's motion with constant acceleration over the prediction horizon is certainly a simplification compared to data-driven prediction systems used in modern autonomous vehicle software. That said, we have found this model to be sufficiently accurate for avoiding vehicles in many instantiations of the head-on scenario \textbf{S3}, particularly when the other vehicle panic brakes after hitting an ice patch (a common reaction of many drivers after losing traction) and ends up simply sliding to a stop.    

We also note that the ESC baseline is a somewhat rudimentary version of stability control in a typical production vehicle, which may include additional features such as sideslip estimation, yaw rate prediction, and active steering interventions. The comparison in this work is meant to highlight the difference in control authority between systems that enable versus those that prevent high sideslip vehicle motion. Future work can incorporate more sophisticated stability control systems to better understand the advantages of drift-capable control. 

\section{Conclusion}
\label{sec:conclusion}
The drift-capable MPC controller presented in this work demonstrates emergent drifting behavior in several winter scenarios based on real-world driving data. These include correcting for oversteer caused by lower friction on the rear axle and maneuvering away from an imminent collision with an oncoming vehicle that has slid directly into its lane. These high sideslip maneuvers emerge naturally from the controller's objective function that encodes safe driving without directly rewarding drifting behavior. Compared with an ESC system, the MPC controller sustains drifts with periods of open-loop instability, demonstrating the tradeoff between stability and controllability often discussed in vehicle dynamics literature. Our Monte Carlo study shows that the drift-capable controller may lose control in scenarios requiring long drifts at the actuator limits, while, within these limits, it performs better than an ESC controller at keeping the vehicle centered in the lane. 

These rare but serious failure cases of MPC reveal new and interesting directions for future work. In particular, real-time feasibility monitoring, perhaps using reachability analysis as in \cite{zhao2022justifying}, could enable the controller to reduce its weights on path or speed tracking in favor of keeping the vehicle in the lane if it detects an unrecoverable drift trajectory. An adaptive cost structure could allow the controller to accept larger lane or speed deviations when actuators are near their limits \cite{zhou2025adaptive}. Finally, implementing this controller as a driver assistance system could give drivers cooperative control of speed, actively balancing risk tolerance and time to destination in real-world settings in which the vehicle is prone to potential traction losses.

\bibliographystyle{IEEEtran}
\bibliography{references} 

\end{document}